\documentclass[]{article}
\usepackage{proceed}
\usepackage{amsmath}
\usepackage{graphicx}
\usepackage{amssymb}
\usepackage{epsfig}
\usepackage{url}
\usepackage{verbatim}

\usepackage{enumitem}

\usepackage{times}
\usepackage{helvet}
\usepackage{courier}
\usepackage{epsfig}
\usepackage{color}

\usepackage{algorithmic}
\usepackage{algorithm}
\usepackage{natbib}
\usepackage{amsfonts}
\usepackage{amsmath,amsthm}
\DeclareMathOperator*{\argmin}{argmin}
\DeclareMathOperator*{\argmax}{argmax}

\newcounter{thm_counter}
\newcounter{lem_counter}

\newcounter{ass_counter}

\newtheorem{theorem}[thm_counter]{Theorem}
\newtheorem{lemma}[lem_counter]{Lemma}

\newtheorem{definition}[ass_counter]{Definition}

\title{Dantzig Selector with an Approximately Optimal Denoising Matrix and its Application to Reinforcement Learning}
\author{
{Bo Liu} \\
{Auburn University}\\
{boliuauburn@gmail.com}
\And
{Luwan Zhang} \\
{University of Wisconsin Madison}\\
{luwanzhang@gmail.com}
\And
{Ji Liu}\\
{University of Rochester}\\
{ji.liu.uwisc@gmail.com}
}

\begin{document} 
\maketitle

\begin{abstract} 
Dantzig Selector (DS) is widely used in compressed sensing and sparse learning for feature selection and sparse signal recovery. Since the DS formulation is essentially a linear programming optimization, many existing linear programming solvers can be simply applied for scaling up. The DS formulation can be explained as a basis pursuit denoising problem, wherein the data matrix (or measurement matrix) is employed as the denoising matrix to eliminate the observation noise. However, we notice that the data matrix may not be the optimal denoising matrix, as shown by a simple counter-example. This motivates us to pursue a better denoising matrix for defining a general DS formulation. We first define the optimal denoising matrix through a minimax optimization, which turns out to be an NP-hard problem. To make the problem computationally tractable, we propose a novel algorithm, termed as ``Optimal'' Denoising Dantzig Selector (ODDS), to approximately estimate the optimal denoising matrix. Empirical experiments validate the proposed method. Finally, a novel sparse reinforcement learning algorithm is formulated by extending the proposed ODDS algorithm to temporal difference learning, and empirical experimental results demonstrate to outperform the conventional ``vanilla'' DS-TD algorithm.
\end{abstract} 

\section{Introduction}
We consider consider the classic problem in compressed sensing, sparse learning, and statistics \citep{donoho2006compressed, candes07dantzig, bickel2009simultaneous}:

\emph{
Given a data (measurement) matrix $X \in {\mathbb{R}^{n \times m}}$ ($m \gg n$) and a noisy observation vector $y \in {\mathbb{R}^n}$ satisfying $y = X\beta^*  + \epsilon $ where $\epsilon$ is the noise vector following the Gaussian distribution $N(0,\sigma^{2}I)$\footnote{The Gaussian distribution can be generalized to any sub-Gaussian distribution.} and $\beta^*$ is the truth model which is a sparse vector. How to recover the sparse vector $\beta^*$ from this under-determined system?
}

Dantzig Selector (DS) \citep{candes07dantzig} is a widely used approach to solving this problem. The standard DS is formulated as
\begin{subequations}
\label{eq:DS}
\begin{align}
\label{eq:DS_obj}
(\text{DS})\quad \hat{\beta}_{\text{DS}} = \argmin_{\beta}~&\|\beta\|_1 \\
\label{eq:DS_cst}
\text{s.t.}~&\|X^T(X\beta-y)\|_\infty \leq \lambda.
\end{align}
\end{subequations}
DS has a very similar performance to another famous formulation LASSO \citep{lasso:tibshirani1996regression} both empirically and theoretically \citep{bickel2009simultaneous}. The DS formulation (\ref{eq:DS}) can be formulated as a linear programming (LP) problem, thus many matured LP solvers 
can be directly applied to address this problem with large-scale problem settings. The DS formulation or its variation has been widely used in reinforcement learning \citep{DantzigRL:2012, ROTD:NIPS2012, mahadevan2012sparse, qin2014sparse}, computational bioinformatics \citep{liu1914statistical}, and computer vision \citep{cong2011sparse}.

The motivation of this paper is to explore the role of $X^T$ (the transpose of $X$) in DS formulation \eqref{eq:DS}. We note that the constraint in \eqref{eq:DS_cst} follows two principles: 1) the defined feasible region should contain the true solution $\beta^*$ with high probability; and 2) to make $\hat{\beta}_{\text{DS}}$ close to $\beta^*$ the feasible region defined by the constraint should be as small as possible, that is, $\lambda$ is expected to a small value. Taking $\beta = \beta^*$ into the constraint leads to the smallest possible value for $\lambda=\|X^T(X\beta^* - y)\|_{\infty} = \|X^T\epsilon\|_{\infty}$.  
If columns of $X$ are normalized to 1, we have $\lim_{n\rightarrow \infty}\|X^T\epsilon\|_\infty \rightarrow 0$ with high probability \citet{candes07dantzig}. Therefore, the factor $X^T$ in the constraint actually plays the role of \textit{denoising}. This motivates us to ask two questions: 1) is $X^T$ the optimal denoising matrix for the recovery of the sparse signal $\beta^*$?; and 2) if not, how to measure the the optimality of the denoising matrix and how to compute the optimal denoising matrix?

Unfortunately, $X^T$ is \emph{not} the optimal denoising matrix in general. We provide a counter-example in Section~\ref{sec:algorithm}.
The main contributions of this paper are summarized below:
\begin{itemize} [noitemsep,topsep=0pt,leftmargin=*]
\item We propose a generalized denoising Dantzig selector formulation (GDDS) and define the optimal denoising matrix for sparse signal recovery via a minimax formulation;
\item A two-stage approach is proposed to  compute the approximately optimal denoising matrix;
\item We apply the proposed ODDS algorithm to an important application in reinforcement learning: the temporal difference learning problem for sparse value function approximation.
\end{itemize}    

This paper is organized as follows: Related work is introduced in Section 2. Section 3, which is the core part of this paper, proposes the general Dantzig Selector formulation with both intuitive motivations as well as the mathematical backgrounds. 
A generalized error bound is proposed, which leads to the definition of the optimal denoising matrix, which is NP-hard in general. To address this problem, a two-stage algorithm for approximately computing the optimal denoising matrix is given. Then in Section 4, the algorithm is applied to  reinforcement learning to design a new algorithm for sparse value function approximation. The experimental results are presented in Section 5, which validate the effectiveness of the proposed algorithm.



\section{Related Work}

The problem considered in this paper has received substantial attentions in compressed sensing, sparse learning, and statistics. The study starts from the special case, i.e.  the noiseless case $\epsilon=0$. To recover the sparse vector $\beta^*$, there are two types of approaches: $\ell_1$ norm minimization and greedy algorithms. The $\ell_1$ norm minimization solves the following optimization problem to estimate $\beta^*$ \citep{bp:chen1994basis, rip:candes2005decoding}
\begin{equation}
\min_{\beta} \quad\|\beta\|_1
\quad
\text{s.t.}\quad X\beta = y.
\end{equation}
The theoretical study \citep{rip:candes2005decoding} suggests that under the restricted isometric property (RIP) condition, the $\ell_1$ norm minimization formulation can recover the true solution $\beta^*$ exactly. The greedy approaches include the forward greedy algorithm (or OMP) \citep{tropp2004greed} and the backward greedy algorithm. A similar theoretical guarantee of exact recovery is proven for the forward greedy algorithm in \citet{zhang2011sparse}.

Now let us turn to the noisy case, i.e.  $\epsilon\neq 0$. The noisy case is more challenging than the noiseless case. It also mainly includes two types of approaches: $\ell_1$ norm minimization approaches and greedy algorithms. To deal with the noise, there are several popular  formulations including DS \citep{candes07dantzig}, LASSO \citep{lasso:tibshirani1996regression}, and basis pursuit denoising (BPDN) \citep{bpdn:chen2001atomic}. They essentially apply different manners to denoise. BPDN uses the $\ell_2$ norm penalty to denoise:
\begin{equation}
\hat{\beta}_{\text{BPDN}} = \argmin_{\beta} \|\beta\|_1
\quad
\text{s.t.}\quad \|X\beta - y\|_2\leq \varepsilon,
\end{equation}
where the constraint basically restricts the noise $\epsilon$ by $\|\epsilon\| \leq \varepsilon$.
DS uses the $\ell_{\infty}$ norm penalty to denoise $\|X^T\epsilon\|_\infty \leq \lambda$ as shown above. LASSO applies the same spirit as DS \citep{bickel2009simultaneous} to denoise, but uses a different formulation
\begin{equation}
\hat{\beta}_{\text{LASSO}} = \argmin_{\beta}{1\over 2}\|X\beta - y\|_2^2 + \lambda \|\beta\|_1 .
\end{equation}
The theoretical error bound for LASSO and DS is similar \citep{bickel2009simultaneous} and better than BPDN in some sense. The key reason lies in that the noise constraint used in DS and LASSO $\|X^T\epsilon\|_\infty \leq \lambda$ is sharper than the noise constraint used in BPDN $\|\epsilon\| \leq \varepsilon$.\footnote{From the optimization perspective, BPDN seems consistent with LASSO, but the theoretical error bound is worse than LASSO and DS for some subtle reasons, which is beyond the scope of this paper.} The greedy approaches mainly include the forward greedy algorithm \citep{zhang2009consistency} and the forward-backward greedy algorithm \citep{zhang2011adaptive, liu2013forward}. 

While Dantzig Selector primarily focuses on $\ell_1$ regularization for sparsity, the group-sparsity structures was explored \citep{ds:group:liu2010group}, and was recently extended to generalized norm \citep{ds:general}. Other variants includes weighted Dantzig Selector \citep{WDS} by re-weighting the sparse signal, multi-stage Dantzig Selector \citep{ds:multi:liu2010multi} for iterative sparse signal recovery, etc. As for the computational achievements, 
besides the primal-dual interior point method \citep{candes07dantzig}, TFOCS \citep{tfocs:becker2011} is also widely used. Later, inexact alternating direction method of multipliers (ADMM) formulations are proposed in \citep{ds:lu2012alternating,ds:wang2012linearized}, which are computationally efficient.

\section{Algorithm}\label{sec:algorithm}
We first show the generic error bound when ``$X^T$'' in \eqref{eq:DS_cst} is substituted by an arbitrary denoising matrix $Q^T$. Then a counter example is provided to show why $X^T$ is not the optimal choice for $Q^T$. We prove an approximate method to pursue the optimal denoising matrix $Q^T$ in the end of this section.

\subsection{Generalized Denoising Dantzig Selector and its Error Bound}

With the backgrounds of the denoising matrix introduced above, one intuitive question is that if $X^T$ the optimal denoising matrix for sparse signal recovery of $\beta^*$? The answer is actually NO! To explain this, we first introduce the generalized denoising Dantzig Selector formulation. Next, an error bound w.r.t the GDDS and regular DS is proposed, and it can be proven that for regular DS, the error bound is tighter than the existing error bound provided in \citep{candes07dantzig} and \citep{bickel2009simultaneous}. Then a simple counter-example is proposed to argue that $X^T$ may not be the optimal denoising matrix.
 We here present some definitions and notations in Figure~\ref{fig:notation}.
\begin{figure}[!ht]
\center
\begin{tabular}{|p{7.5cm}|}
\hline \vskip 0.01in
\begin{itemize} [noitemsep,topsep=0pt,leftmargin=*]
\item $h_T$,  $h_{T^c}$: $T$ and $T^c$ are two complementary subsets in $\{1,2,\cdots, m\}$. We denote by $h_{T}\in \mathbb{R}^{m}$ the vector taking the same values as $h$ on $T$ and zeros in the rest; the same for $h_{T^c} \in \mathbb{R}^m$; 
\item $|T|$ returns the cardinality of the set $T$;
\item $\|W\|_{\infty}$ is defined as the induced $\infty$ norm of matrix $W \in \mathbb{R}^ {m \times m}$, i.e.  $\|W\|_{\infty} = \max_{x} {\|W x\|_{\infty} \over \|x\|_\infty}$ where the vector infinity norm $\|W x\|_{\infty}$ and $\|x\|_\infty$ are defined as usual; 
\item $\leq_{\text{(P)}}$: less than or equal to with high probability;
\end{itemize}
\\\hline
\end{tabular}
\caption{Notation used in this paper}
\label{fig:notation}
\end{figure}
 
First we define a generalized denoising Dantzig Selector (GDDS) formulation by using a general denoising matrix $Q^T$ ($Q\in \mathbb{R}^{n\times m}$) to replace the $X^T$ in the constraint: 
\begin{equation}
\begin{aligned}
(\text{GDDS})\quad \min_{\beta}:~&\|\beta\|_1 \\
\text{s.t.}:~&\|Q^T(X\beta-y)\|_\infty \leq \lambda.
\label{eq:GDDS}
\end{aligned}
\end{equation}

Next we will propose an error bound for the proposed GDDS in \eqref{eq:GDDS} and regular DS.
Since the denoising matrix is not necessarily $X^T$ anymore, the commonly used RIP condition or restricted eigenvalue (RE) condition     \citep{van2009conditions} are not eligible here. To extend to the general case, we define a new condition termed as \textit{generalized restricted (GR) constant}.
\begin{definition}\label{ass:1}
(GR constant) Given $X, Q\in \mathbb{R}^{n \times m}$ and $p\in [1,\infty]$, the general restricted constant $\rho (Q^TX,p,s)$ is defined as
\begin{equation}
\rho (Q, X,p,s) := \min_{|T|\leq s, \|h_{T^c}\|_1\leq \|h_T\|_1}
\frac{\|Q^TXh\|_\infty}{\|h\|_p}.
\label{eq:kappa}
\end{equation}
\end{definition} 
{ This definition essentially provides the lower bound of the ratio between $\|Q^TXh\|_{\infty}$ and $\|h\|_p$ over $h$ in a subset of $\mathbb{R}^{p}$, which characterizes the property of $Q^TX$. The GR constant leads to a weaker  condition to exactly recover the sparse signal $\beta^*$ for the noiseless case ($\epsilon = 0$) and a tighter error bound for the noisy case ($\epsilon \neq 0$) than the existing analysis, as shown by Theorem~\ref{thm:bound}.} 
Based on the definition for GR constant in Definition~\ref{ass:1}, we have the following error bound on $\|\hat{\beta}_{\text{GDDS}} - \beta^*\|_p$. 
\begin{theorem} \label{thm:bound}
Assume that the GR condition is satisfied, i.e. the GR constant $\rho(Q,X, p, \|\beta^*\|_0) > 0$. Choose $\lambda = \|Q^T\epsilon\|_{\infty}$ in \eqref{eq:GDDS}. We have 
\begin{equation}
\|\hat{\beta}_{\text{GDDS}} - \beta^*\|_p \leq \frac{2\|Q^T\epsilon\|_{\infty}}{\rho(Q,X,p, \|\beta^*\|_0)},
\label{eq:bound}
\end{equation}
where $\hat{\beta}_{\text{GDDS}}$ is the solution to \eqref{eq:GDDS} and $p$ can be any value in the range $[1, \infty]$.
\end{theorem}


It should be noted that albeit with a more general error bound, Theorem~\ref{thm:bound} does not weaken the existing analysis based on the following two observations:
\begin{itemize}[noitemsep,topsep=0pt,leftmargin=*]
\item In the noiseless case, i.e.  $\epsilon = 0$, if the GR condition is satisfied, then $\hat{\beta}_{\text{GDDS}}$ is able to exactly recover $\beta^*$. Note that the GR condition is weaker than the RIP condition for $Q=X$ \citep{candes2008restricted}. In other words, the RIP condition leads to GR condition when $Q=X$. The detailed interpretation is provided in Appendix.
\item In the noisy case, i.e.  $\epsilon \neq 0$, this bound \eqref{eq:bound} is a tighter bound than the bound $\|\hat{\beta}_{\text{DS}} - \beta^*\|_p$ for DS in \citep{candes07dantzig, bickel2009simultaneous} for $Q=X$. Please also refer to Appendix for detailed comparisons.
\end{itemize}

The key reason why Theorem~\ref{thm:bound} does not loose the existing analysis lies in that the definition of the GR constant $\rho(Q,X, p, s)$ skips many relaxation steps by directly indicating the relationship between $\|Q^T\epsilon\|_{\infty}$ and $\|\hat{\beta}_{\text{GDDS}} - \beta^*\|^2$. 
Given $p$, $s$ and $X$, $\rho(Q,X,p,s)$ reflects the ability for sparse signal recovery of $\beta^*$ of the denoising matrix $Q^T$: \emph{The larger $\rho(Q,X,p,s)$ is, the better $Q$ is able to recover the sparse signal $\beta^*$. }
Therefore,  since the error bound provided in Theorem is tight enough, it is reasonable to use the bound in \eqref{eq:bound} as the evaluation criteria to see why $X^T$ is not the optimal denoising matrix and define the optimal denoising matrix. 

\subsection{A Counter-Example}
 To see why $X^T$ may not be the optimal denoising matrix, we show an example of $Q$ which gives a lower value for the bound in \eqref{eq:bound}. To construct such a matrix $Q$, we require the same conditions on $Q$ as $X$, i.e.  all columns of $Q$ have been normalized with norm $1$. We can verify that any $Q$ with unit column norm, the value of $\|Q^T\epsilon\|_{\infty}$ is comparable to $\|X^T\epsilon\|_{\infty}$, according to the following standard results in Lemma~\ref{lem:1}.
\begin{lemma} \label{lem:1}
 For any $Q\in \mathbb{R}^{n\times m}$ satisfying $\|Q_{.i}\|=1$ for $i=1,\cdots,m$, we have 
 \[
 \|Q^T\epsilon\|_\infty \leq \sigma\sqrt{\log m}
 \]
with high probability at least $1 - O(1/m)$.
\end{lemma}


The key reason why $\|Q^T\epsilon\|_{\infty}$ and $\|X^T\epsilon\|_{\infty}$ are comparable lies in that all entries in random vectors $Q^T\epsilon \in \mathbb{R}^{p}$ and $X^T\epsilon \in \mathbb{R}^p$ follow the same Gaussian distribution $\mathcal{N}(0, \sigma^2)$. 
{ Since $\|Q^T\epsilon\|_\infty$ and $\|X^T \epsilon\|_{\infty}$ are comparable, we only need to find an example for $Q$ such that $\rho(Q,X, p, s)>\rho(X,X, p, s)$ for some $X$.}  Let $s=1$ for simplicity and
\begin{equation}
X = \left(
           \begin{array}{cc}
             \sqrt{3}/2 & 1/2 \\
             1/2  &\sqrt{3}/2 \\
           \end{array}
         \right),~
Q=\left(
           \begin{array}{cc}
             1 & 0 \\
             0 & 1 \\
           \end{array}
         \right).
\end{equation}
We have
\begin{equation}
X^TX = \left(
           \begin{array}{cc}
             1 & \sqrt{3}/2 \\
             \sqrt{3}/2 & 1 \\
           \end{array}
         \right)
\end{equation}
\begin{align*}
\rho(X,X,p,1) = &\min_{\stackrel{|T|\leq 1}{\|h_{T^c}\|_1\leq \|h_T\|_1}} \frac{\|X^TXh\|_\infty}
{\|h_T\|_p} \\
= & 1-\sqrt{3}/2 = 0.134,\\
\rho(Q,X,p,1) = &\min_{\stackrel{|T|\leq 1}{\|h_{T^c}\|_1\leq \|h_T\|_1}} \frac{\|Q^TXh\|_\infty} {\|h_T\|_p} \\
= &\sqrt{3}/2 - 1/2 = 0.366,
\end{align*}
where in both cases the optimal value of $h$  is $h = [1, -1]^T$ regardless of the value of $p$. Thus, this example shows that $X^T$ may not always be the optimal denoising matrix.

\subsection{Optimal Denoising Matrix and its Approximation}

From the counter-example in the previous section, we know that $X$ is not the optimal option for $Q$ to maximize $\rho(Q,X, p, s)$. Therefore, it raises an optimization problem to find the optimal $Q$:
\begin{equation}
\begin{aligned}
Q^* = \argmax_{\|Q_{.i}\|\leq 1}\min_{\stackrel{|T|\leq s}{\|h_{T^c}\|_1\leq \|h_T\|_1}} \frac{\|Q^TXh\|_\infty} {\|h_T\|_p}.
\label{eqn_org}
\end{aligned}
\end{equation}


However, this formulation is extremely difficult to solve. Although we can solve it easily for small $n,m$ as in our example above, it is NP-hard in general. Therefore, it is unrealistic to solve this problem exactly. To find a reasonable approximation, we consider an alternative way to finding an optimal matrix $W\approx Q^TX$:
\begin{equation}
W^* = \argmax_{\|W\|_{\infty} \leq 1}\min_{\stackrel{|T|\leq s}{\|h_{T^c}\|_1\leq \|h_T\|_1}} \frac{\|Wh\|_\infty} {\|h_T\|_p}.
\end{equation}
 
One can easily verify that the optimal solution is $W^* = I$.\footnote{Actually, it is not important which norm is used to restrict $W$. For most norms we verified, the optimal solution for $W$ should be a diagonal matrix with equal values.} 
The second step is to find the optimal $Q$. We try to find the best $Q$ from another perspective. Intuitively, we want $Q^TX$ to be close to the identity matrix. We estimate $Q$ by the following:
\begin{equation}
\begin{aligned}
\min_{Q}:~&\|Q^TX-I\|_{p}\quad
\text{s.t.}:~&\|Q_{.i}\|\leq 1
\end{aligned}
\label{eq:q_original}
\end{equation}
where $p\in [1, \infty]$ and $\|X\|_p:= (\sum_{i,j}|X_{ij}|^p)^{1/p}$. We have many options for choosing $p\in [1, \infty]$. Based on our empirical study, a reasonable empirical value for $p$ is $p=2$, and thus the problem can be recast as a strongly convex problem
\begin{equation}
\begin{aligned}
\min_{Q}:~& \|Q^TX-I\|^2_2\quad
\text{s.t.}:~&\|Q_{.i}\|\leq 1
\end{aligned}
\label{eq:q}
\end{equation} 
There are many optimization algorithms to address this problem. We  use Nesterov's accelerated gradient method \citep{nesterov2004introductory}. Note that the empirical $Q$ obtained from \eqref{eq:q_original} or \eqref{eq:q} is generally different from the optimal one defined in \eqref{eqn_org}.
And the optimal denoising Dantzig Selector algorithm is summarized as in Algorithm 1.
\begin{algorithm}
\label{alg:ODDS}
\caption{``Optimal'' Denoising Dantzig Selector (ODDS)}
\begin{algorithmic}
\REQUIRE $X \in \mathbb{R}^{n\times m}, y \in \mathbb{R}^{n}$\\
\ENSURE $\beta$
\STATE Compute the denoising matrix $Q^T$ via Eq. \eqref{eq:q} \\
\STATE Compute $\beta$ via Eq.~\eqref{eq:GDDS}
\end{algorithmic}
\end{algorithm}

\begin{figure}
\vspace{-10mm}
\includegraphics[width=0.4\textwidth]{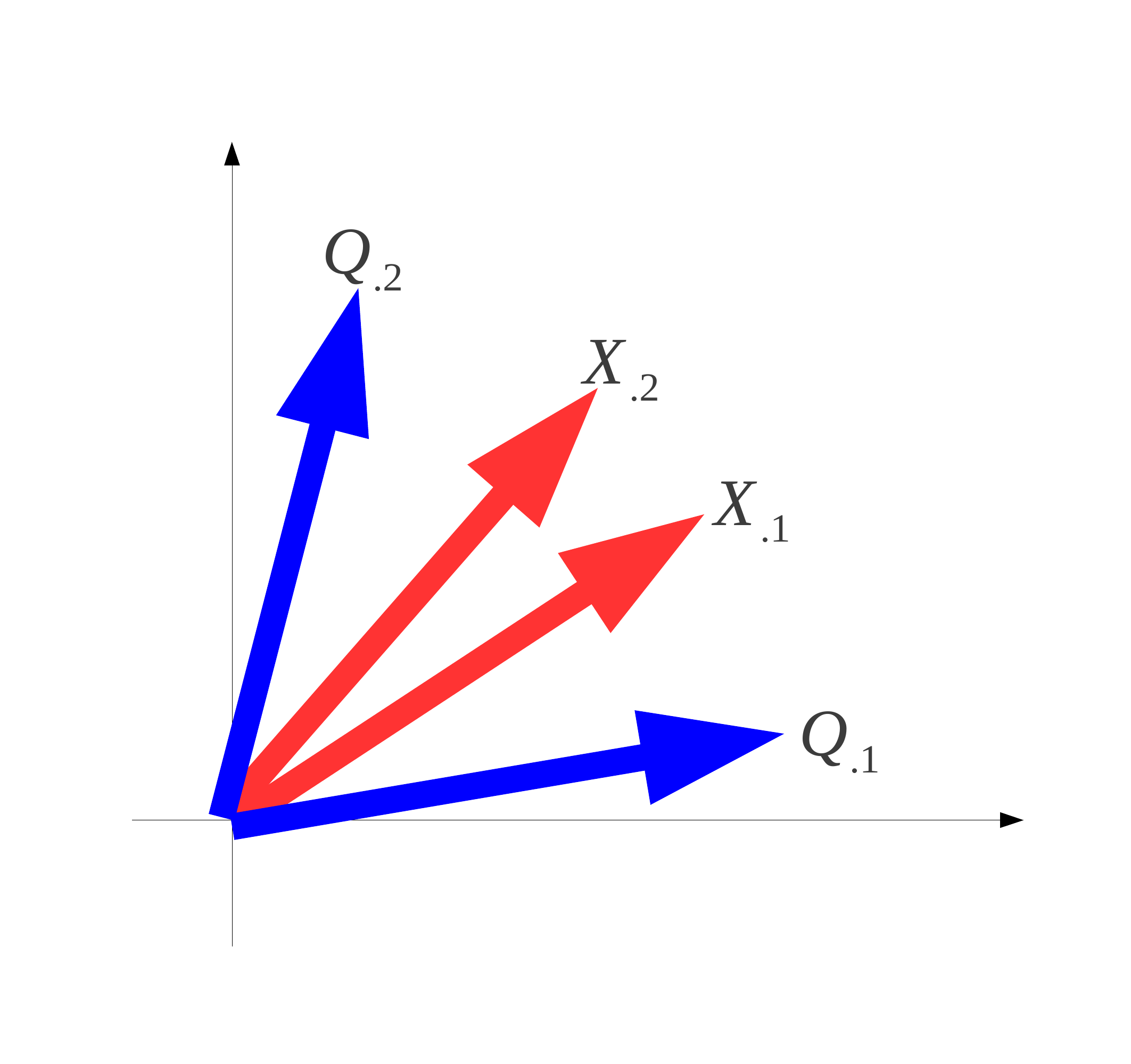}
\vspace{-10mm}
\caption{An example of $X$ and $Q$.}
\label{fig:example}
\end{figure}

Figure~\ref{fig:example} provides an example of $Q$ and $X$ in the two-dimensional case. Intuitively, $Q$ is an approximation to $X$ by slightly modifying all feature (column) vectors of $X$ such that they are as different from each other as possible.

\section{Reinforcement Learning \label{sec:Background}}
Dantzig selector has an important application in reinforcement learning. This section first briefly introduces reinforcement learning and then shows how to apply the proposed ODDS method to it.

A {\em Markov Decision Process} (MDP) is defined by the tuple $({S,A,P_{ss'}^{a},R,\gamma})$,
comprised of a set of states $S$, a set of actions $A$, a dynamical system model comprised of the
transition kernel $P_{ss'}^{a}$ specifying the probability of transition
 from state $s \in S$ to state $s' \in S$ under action $a \in A$, a reward model $R(s,a):S \times A \to \mathbb{R}$,
and $0\leq\gamma<1$ is a discount factor. A policy $\pi:S\rightarrow A$
is a deterministic mapping from states to actions. Associated with
each policy $\pi$ is a value function $V^{\pi}$, which is the fixed
point of the Bellman equation: \vskip -0.5cm 
\[
{V^\pi } = {T^\pi }({V^\pi }) = {R^\pi } + \gamma {P^\pi }{V^\pi },
\]
\vskip -0.4cm 
for a given state $s \in S$, ${R^\pi }(s) = R(s,\pi (s))$, and $P^{\pi}$ is the state transition function
under fixed policy $\pi$, and $T^{\pi}$ is known as the \emph{Bellman
operator}. In what follows, we often drop the dependence of $V^{\pi},T^{\pi},R^{\pi}$
on $\pi$, for notation simplicity. In linear value function approximation,
a value function is assumed to lie in the linear span of a basis function
matrix $\Phi$ of dimension $\left|S\right|\times d$, where $d$
is the number of linear independent features.  
It is easy to show that the ``best'' approximation ${\hat v_{{\rm{best}}}}$ of the true value function $V$   satisfies the following equation,  
\begin{equation}
{\hat v_{{\rm{best}}}} = \Pi V = \Pi {L^{ - 1}}R,
\end{equation}
where $\Pi  = \Phi {({\Phi ^T}\Xi \Phi )^{ - 1}}{\Phi ^T}\Xi $, 
$\Xi$ is a diagonal matrix where the $i$-th diagonal entry $\xi_i$ is the stationary state distribution w.r.t state $s_i$.
and $L=I-\gamma P$. However, ${L^{ - 1}}$ is also computationally prohibitive. To this end, a more practical way is to compute a Galerkin-Bubnov approximate solution \citep{yu2010error} is via solving a fixed-point equation 
\begin{equation}
\hat v = \Pi _{_\Phi }^XT\hat v
\label{eq:obfp}
\end{equation}
 w.r.t an oblique projection    $\Pi _{_\Phi }^X$ onto ${\rm span}(\Phi)$ orthogonal to $X$, i.e., $\Pi _{_\Phi }^X = \Phi {({X^T}\Phi )^{ - 1}}{X^T}$. 
 The existence and uniqueness of the solution can be verified since $T$ is a contraction mapping, $\Pi _{_\Phi }^X$ is a non-expansive mapping, and thus $\Pi _{_\Phi }^X T$ is a contraction mapping. 
 As for the optimal solution $\hat v_{\rm best}$, we have ${X_{{\rm{best}}}} = {({L^T})^{ - 1}}\Xi \Phi $, which is also computational expensive.
 An often used $X$ used in the fixed-point equation (\ref{eq:obfp}) is ${X_{TD}} = \Xi \Phi $, which is computationally affordable, and the corresponding solution to the fixed-point equation is ${{\hat v}_{TD}}$.
 The other often used $X$ is ${X_{BR}} = \Xi L\Phi $ \citep{Scherrer:ObliqueProjection}, which we will not explained in details here.
 However, it is obvious that none of ${X_{TD}}$ or ${X_{BR}}$ is the optimal solution. 
 In fact, it has been shown that $||V - {{\hat v}_{TD}}|{|_\Xi } \le \frac{1}{{1 - \gamma }}||V - {{\hat v}_{{\rm{best}}}}|{|_\Xi }$ \citep{tsitsiklis-roy:tdfun}, which implies that approximation error $||V - {{\hat v}_{TD}}|{|_\Xi }$ between the true value function $V$ and ${{\hat v}_{TD}}$ can be arbitrarily bad when $\gamma  \to 1$. 
Given a fixed-point equation (\ref{eq:obfp}), we have
\begin{equation}
\Pi _\Phi ^XT\hat v - \hat v = \Pi _\Phi ^X(T\hat v - \hat v) = \Phi {({X^T}\Phi )^{ - 1}}{X^T}(T\hat v - \hat v)
\end{equation}
Thus we can formulate the approximation error via constraining
$||{X^T}(T\hat v - \hat v)|{|_\infty } \le \lambda $. 

Since the $P, R, \Xi$ models are not accessible, the sample-based estimation is used.
Given $n$ training samples of the form $\big(s_i,a_i,r_i,s'_i\big)_{i=1}^n,\;s_i\sim\Xi,\;r_i=r(s_i,a_i),\; a_i\sim\pi_b(\cdot|s_i),\;s'_i\sim P(\cdot|s_i,a_i)$, the  empirical Bellman operator $\hat{T}$ is thus written as
\begin{equation}
\label{eq:EmpBellmanEq}
\hat{T}(\hat \Phi \theta ) = \hat R + \gamma\hat\Phi '\theta
\end{equation}
We denote $\hat{\Phi}$ (resp. $\hat{\Phi}'$) the empirical feature matrices whose $i$-th row is the feature vector $\phi {({s_i})^T}$ (resp. $\phi {({s'_i})^T}$), and $\hat{R} \in  \mathbb{R}^n$ the empirical reward vector with corresponding  $r_i$ as the $i$-th row. 
And the error constraint can be formulated as $||{Q^T}(A\theta  - b)|{|_\infty } \le \lambda $, where $A = \hat \Phi  - \gamma \hat \Phi ',b = \hat R$, and $Q \in {\mathbb{R}^{n \times d}}$.  
So given $Q$ and a pre-defined error $\lambda$, the Dantzig Selector temporal difference learning problem can be formulated as
\begin{equation*}
\hat{\theta}_{\rm{DS-TD}} = \argmin_\theta  \|\theta \|_1\quad \text{s.t.}\quad \|{Q^T}({A}\theta  - {b})\|_\infty \le \lambda.
\end{equation*} 
If $Q={\hat \Phi }$, then the algorithm is the DS-TD algorithm \citep{DantzigRL:2012}. 
 Motivated by the GDDS algorithm to find the optimal denoising matrix $Q^T$, we design a new Dantzig Selector temporal difference learning algorithm to compute the optimal $Q^T$ for sparse value function approximation as follows
\begin{equation}
\hat{\theta}_{\rm{ODDS-TD}} = \argmin_\theta  \|\theta \|_1\quad \text{s.t.}\quad \|{Q^T}(A\theta  - b)\|_\infty \le \lambda
\label{eq:odstd}
\end{equation}
where $Q$ is computed from solving
\begin{equation}
\begin{aligned}
\min_{Q}:~&\|Q^TA-I\|^2_F\quad
\text{s.t.}:~&\|Q_{.i}\|\leq 1
\end{aligned}
\label{eq:q-td}
\end{equation}
Based on above, we propose the following ODDS-TD algorithm. 
\begin{algorithm}
\label{alg:ods-td}
\caption{``Optimal'' Denoising Dantzig Selector for Temporal Difference Learning with  (ODDS-TD)}
\begin{algorithmic}
\REQUIRE $A, b$
\ENSURE $\hat{\theta}_{\rm{ODDS-TD}}$
\STATE Compute the denoising matrix $Q^T$ via Eq. (\ref{eq:q-td})
\STATE Compute $\hat{\theta}_{\rm{ODDS-TD}}$ via Eq. (\ref{eq:odstd})
\end{algorithmic}
\end{algorithm}
 \vspace{-3mm}

\section{Empirical Experiment}
Experiments are first conducted on synthetic data with small-scale size and medium-scale respectively. Next, we apply the proposed method to a  reinforcement learning problem on real data sets to show its advantage over existing algorithms.


\subsection{Small-Scale Experiment}
\vspace{-3mm}

\begin{figure}[t]
\vspace{-5mm}
\includegraphics[width=0.5\textwidth]{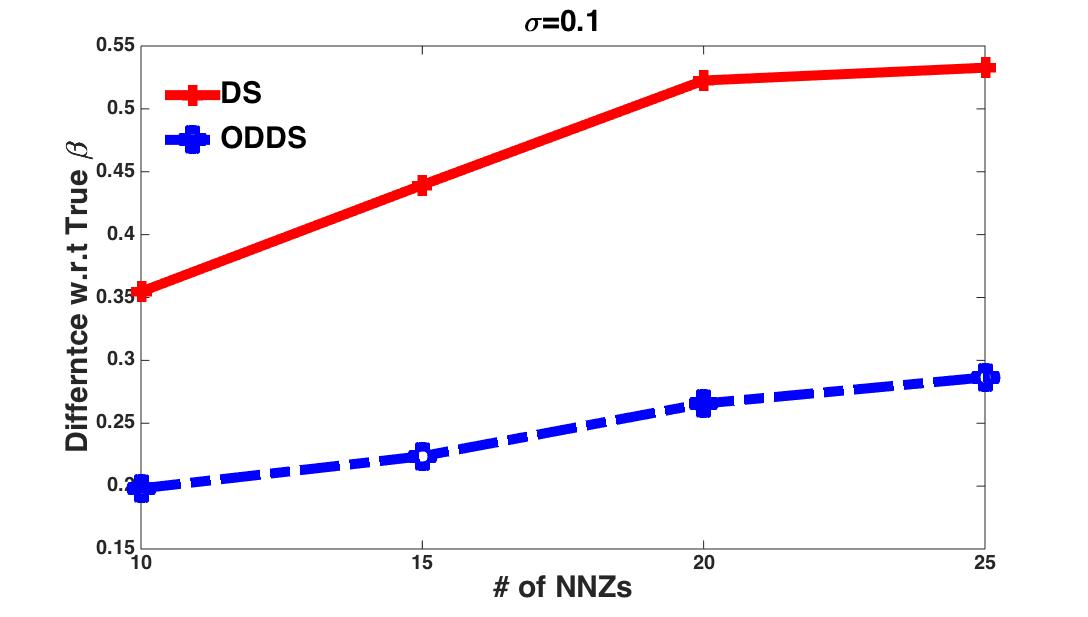}\\
\includegraphics[width=0.5\textwidth]{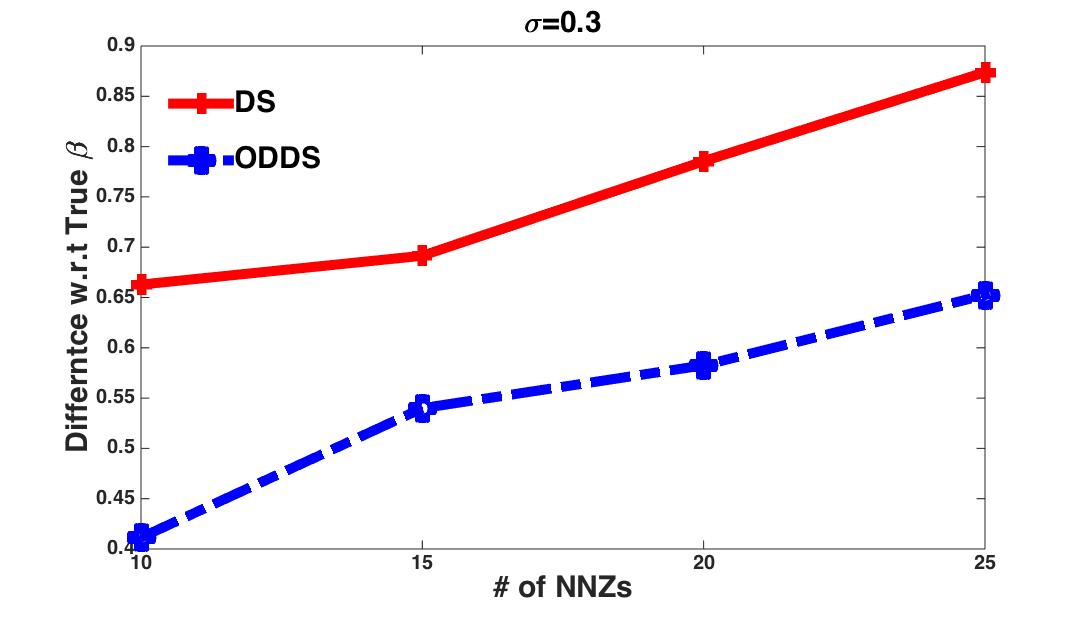}\\
\includegraphics[width=0.5\textwidth]{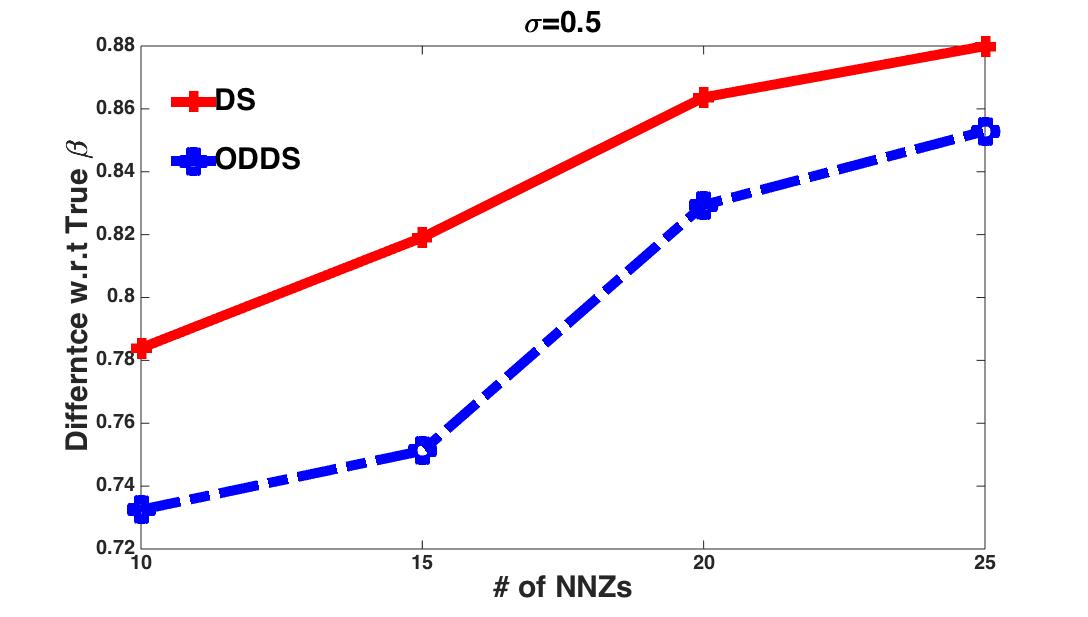}
\vspace{-5mm}
\caption{Small-scale Experiment w.r.t different sparsity levels.}
\vspace{-5mm}
\label{fig:small}
\end{figure}

In this experiment, we conduct the comparison study between the regular Dantzig Selector (DS) and ODDS. We first compare the performance of
different algorithms w.r.t different sparsity and noise levels. We set $n=100,m=150$, the true signal $\beta^*$ is preset, with different numbers of non-zero elements (NNZ) among $10,15,20,25$. 
We also change different noise level varying among $\sigma=0.1,0.2,0.3$. 
Figure~\ref{fig:small} shows the performance comparison of the two algorithms, wherein in each subfigure, the noise level is the same, and the $x$-axis denotes the NNZ level, and the result is measured by the difference of the learned sparse signal with $\beta^*$. From Figure~\ref{fig:small}, we can see that $||{\hat \beta _{{\rm{ODDS}}}} - {\beta ^*}|{|_2}$ is much smaller than $||{\hat \beta _{{\rm{DS}}}} - {\beta ^*}|{|_2}$.

\subsection{Medium-Scale Experiment}
\vspace{-3mm}

\begin{figure}[t]
\vspace{-5mm}
\includegraphics[width=0.5\textwidth]{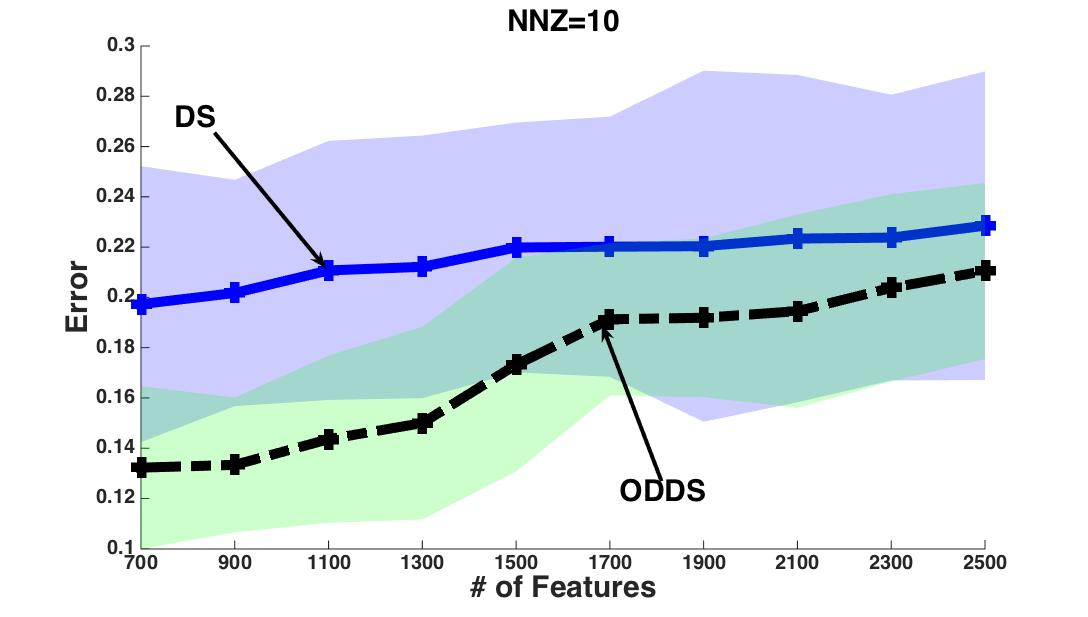}\\
\includegraphics[width=0.5\textwidth]{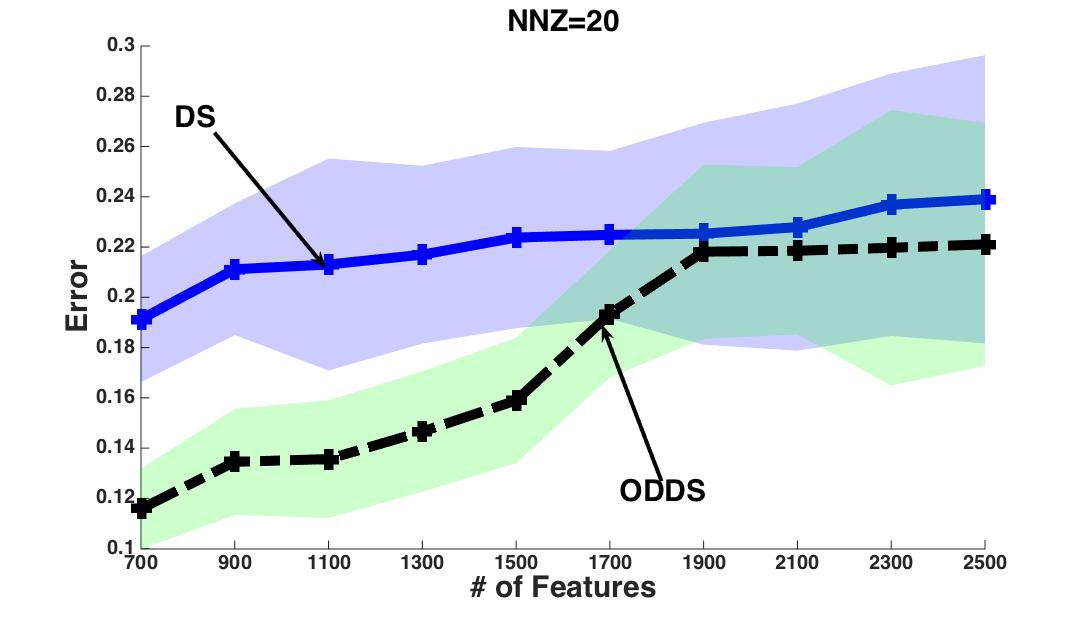}\\
\includegraphics[width=0.5\textwidth]{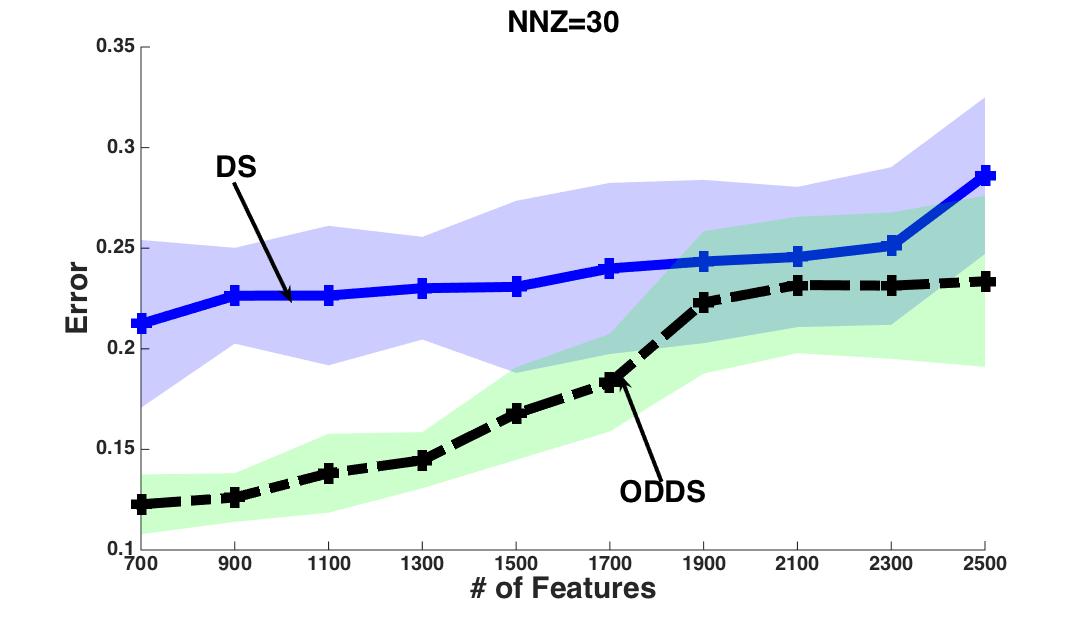}
\vspace{-5mm}
\caption{Comparison w.r.t different number of features.}
\vspace{-5mm}
\label{fig:mid}
\end{figure}

In this experiment, we conduct the comparison study between the regular
DS and ODDS. The experimental setting is that given $n=500$, the number
of features $m$ goes among $700,900,\cdots,2500$, and $NNZ = 10$. The noise level $\sigma=0.01$,
and $\lambda$ is chosen as $\lambda=\sigma\sqrt{2\log n}$ as suggested by \citep{candes07dantzig} 
for a fair comparison between DS and ODDS, and the result is averaged by the mean-squares error (MSE), which is averaged over
$50$ runs. From Figure~\ref{fig:mid}, we can see that the performance
of ODDS is much better than that of regular DS, with both much lower MSE
error and less variance.

\subsection{ODDS-TD Experiment}
\vspace{-3mm}

\begin{figure}[t]
\vspace{-3mm}
\includegraphics[width=0.5\textwidth]{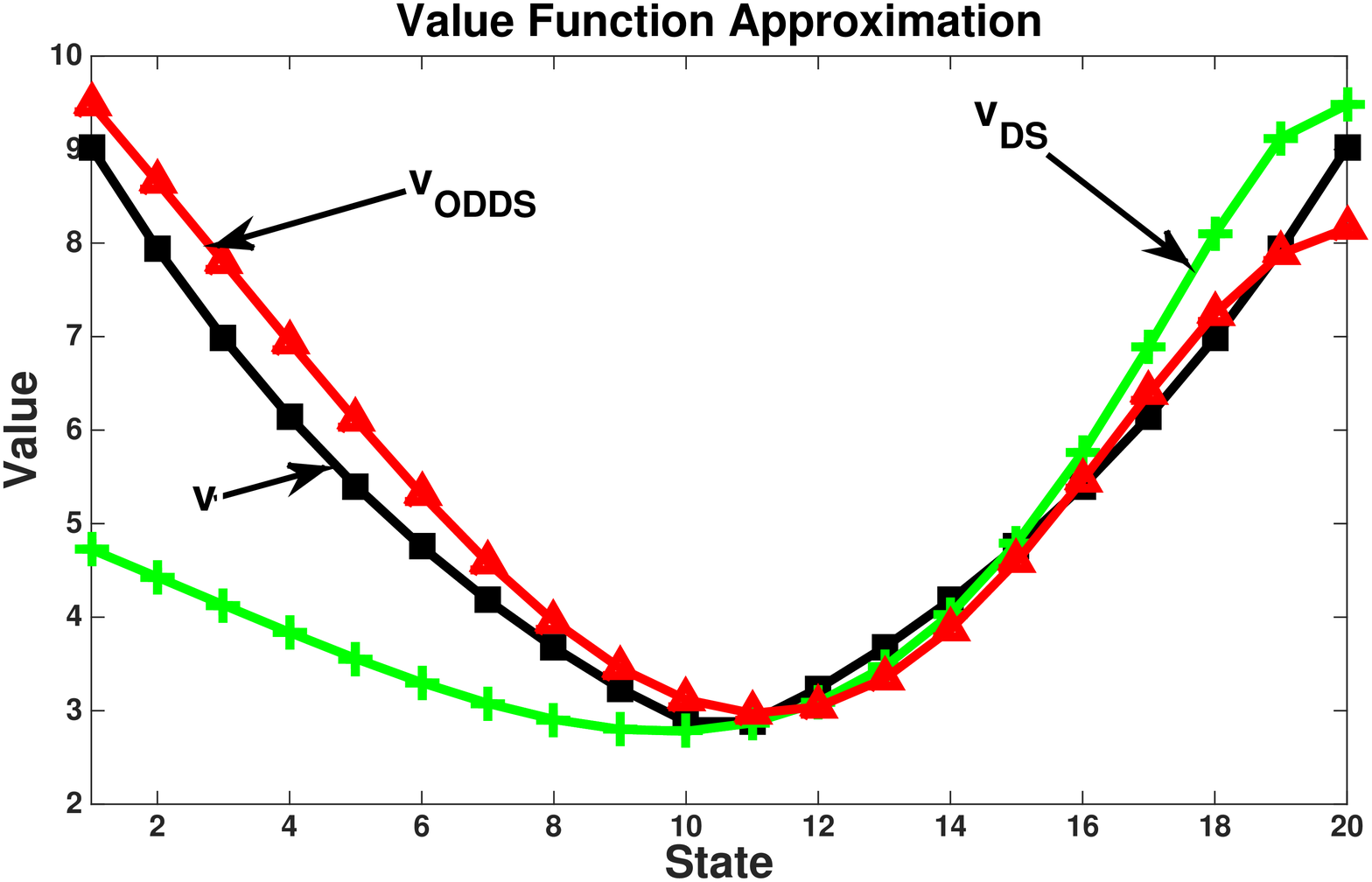}\\
\includegraphics[width=0.5\textwidth]{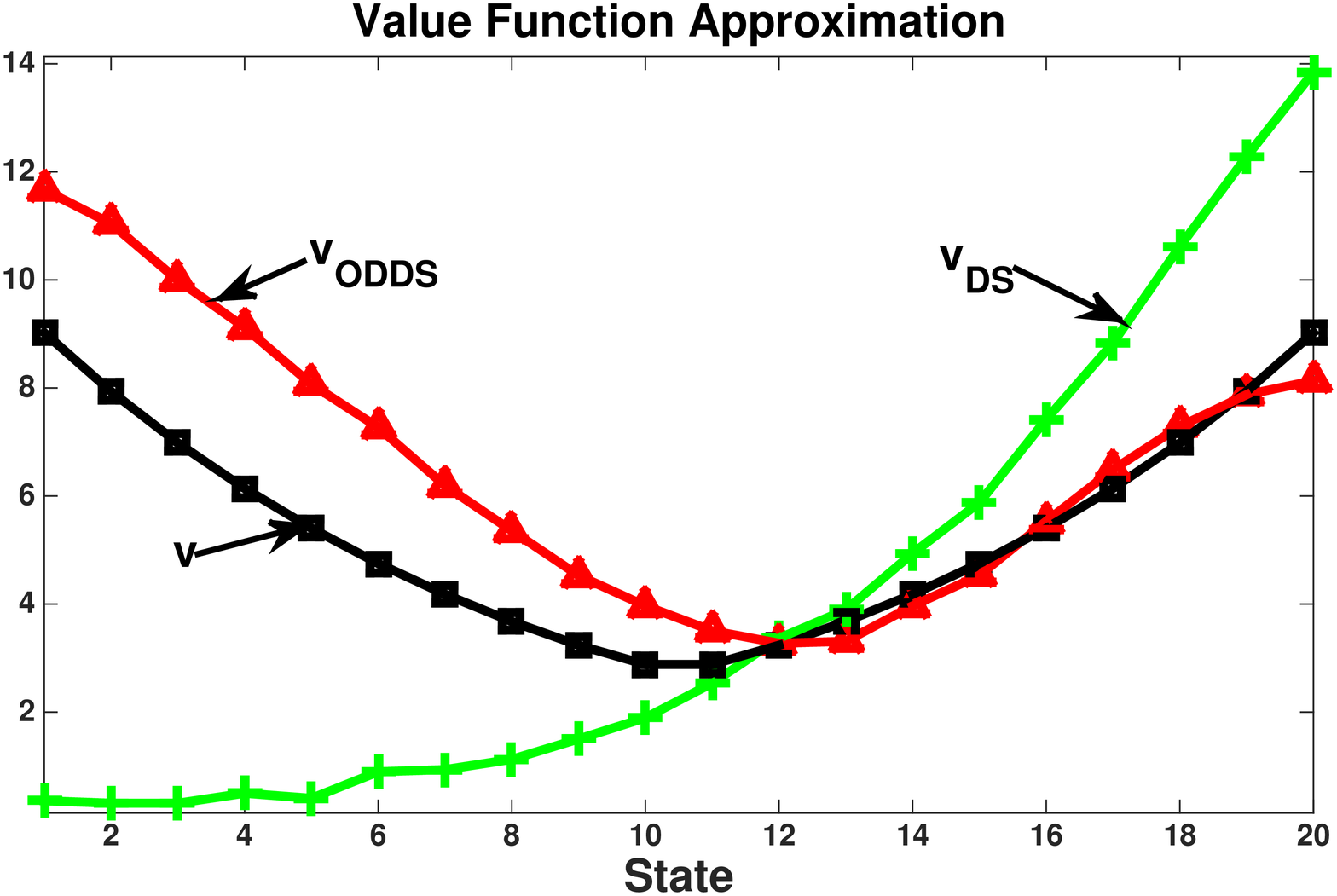}
\vspace{-7mm}
\caption{Comparison between ODDS-TD and DS-TD}
\vspace{-5mm}
\label{fig:ODDStd}
\end{figure}

In this experiment, we compare the performance of the DS-TD \citep{DantzigRL:2012} and the proposed ODDS-TD algorithm. We test the
performances of the two algorithms on the $20$-state corrupted chain
domain, wherein the two goal-states are $s\in[1,20]$ with
the reward signal $+1$, and the probability transition to the nearest
goal-state is $0.9.$ The features are constructed as follows. $5$
radial basis functions (RBF) are constructed, and one constant is
used as an offset, and all other noisy features are randomly
drawn from Gaussian distribution. So for each sample $s_{t}$, the
feature vector $\phi(s)=[1,{\rm {RBF}}(1),\cdots,{\rm {RBF}}(5),{s_{1}},\cdots,{s_{n}}].$  
 $200$ off-policy samples are collected via randomly sampling the
state space. Two comparison studies are carried out. In the first experiment,
there are $300$ noisy features, so altogether there
are $306$ features, and the value function approximation result is
shown in the first subfigure of Figure~\ref{fig:ODDStd}, where $v_{ODDS}$ and $v_{DS}$
are the value function approximation results of ODDS and regular DS. From the figures, one can see that the $v_{ODDS}$ is much more accurate
than that of $v_{DS}$. The second experiment poses an even more challenging
task, where the number of noisy features is set to be $600$,
so altogether there are $606$ features, which makes the feature selection
task much more difficult, and the result is shown in the second subfigure of Figure~\ref{fig:ODDStd}. We
can see that $v_{DS}$ has been severely distorted, whereas $v_{ODDS}$
is still able to well preserve the topology of the value function,
and is able to generate the right policy.

\section{Conclusion}
\vspace{-3mm}

In this paper, motivated by achieving a better sparse signal recovery, we propose a generalized denoising matrix Dantzig selector formulation. A two-stage algorithm is proposed to find the optimal denoising matrix. The algorithm is then applied to sparse value function approximation problem in temporal difference learning field, and the empirical results validate the efficacy of the proposed algorithm.
There are many interesting future directions along this research topic. For example, the ODDS framework can be extended to weighted Dantzig Selector (WDS) \citep{WDS}, multi-stage Dantzig Selector \citep{ds:multi:liu2010multi}, group Dantzig Selector \citep{ds:group:liu2010group} and generalized Dantzig Selector \citep{ds:general}, which are parallel research directions of the Dantzig Selector algorithms family.

\section{Acknowledgements}
Bo Liu is partially funded by NSF Grant IIS-1216467, and the startup funding at Auburn University. Ji Liu is funded by NSF Grant CNS-1548078, the NEC fellowship, and the startup funding at the University of Rochester.

\appendix
\section{Theoretical Analysis of ODDS}
\noindent{\bf Proof to Theorem~\ref{thm:bound}}
\vspace{-.5cm}
\begin{proof}
Denote the difference between the solution to \eqref{eq:GDDS} $\hat{\beta}_{\text{GDDS}}$ and the true model $\beta^*$ as $h=\hat{\beta}_{\text{GDDS}} - \beta^*$. Denote the support set of $\beta^*$ by $T \subset \{1,2,\cdots, m\}$. First we verify that the true model $\beta^*$ is a feasible point to the problem~\eqref{eq:GDDS} due to the following observation
\[
\|Q^T(X\beta^* - y)\|_{\infty} = \|Q^T\epsilon\|_{\infty} \leq \lambda.
\]
Since $\hat{\beta}_{\text{GDDS}}$ is the minimizer of problem ~(\ref{eq:GDDS}), it follows that
\begin{align*}
& \|\hat{\beta}_{\text{GDDS}}\|_1 \leq \|\beta^*\|_1 
\\ \Rightarrow & \sum_{j \in T} |(\hat{\beta}_{\text{GDDS}})_j| + \sum_{j \in T^c} |(\hat{\beta}_{\text{GDDS}})_j| \leq \sum_{j \in T} |\beta^*_j|
\\ \Rightarrow &
\sum_{j \in T^c} |(\hat{\beta}_{\text{GDDS}})_j| \leq \sum_{j \in T} |\beta^*_j - (\hat{\beta}_{\text{GDDS}})_j|
\\ \Rightarrow &
\|h_{T^c}\|_1 \leq \|h_T\|_1.
\end{align*}
From the definition of $\rho(Q,X, p, s)$ in \eqref{eq:kappa}, we have
\[
\rho(Q,X, p, \|\beta^*\|_0) \|h\|_p \leq \|Q^TXh\|_{\infty}
\]
which indicates 
\begin{equation}
\|h\|_p \leq \frac{\|Q^TXh\|_{\infty}}{\rho(Q,X, p, \|\beta^*\|_0)}.
\label{eq:proof_thm}
\end{equation}
It follows that 
\begin{align*}
\|Q^TXh\|_{\infty} = & \|Q^TX(\hat{\beta}_{\text{GDDS}} - \beta^*)\|_{\infty}  
\\ 
= & \|Q^T(X\hat{\beta}_{\text{GDDS}} - y + \epsilon)\|_{\infty}
\\
\leq &  \|Q^T(X\hat{\beta}_{\text{GDDS}} - y)\|_{\infty} +  \|Q^T \epsilon\|_{\infty}
\\ 
\leq & 2\|Q^T\epsilon\|_{\infty}.
\end{align*}
Combining \eqref{eq:proof_thm}, we obtain the desired error bound
\[
\|h\|_p \leq \frac{2\|Q^T\epsilon\|_{\infty}}{\rho(Q,X, p, \|\beta^*\|_0)}.
\]
It completes the proof.
\end{proof}

{
\noindent{\bf Proof of Lemma~\ref{lem:1}}
\begin{proof}
This proof follows the standard union bound proof. For completion, we provide the proof below. Since $\epsilon \sim \mathcal{N}(0, I_{n\times n}\sigma^2)$, for each linear combination of $\epsilon$, we have
$Q_{.i}^T\epsilon \sim \mathcal{N}(0,\sigma^2)$. Recall that for any $t>0$, we have that
\begin{equation}
\int_t ^\infty e^{-x^2/2} dx \leq e^{-t^2/2}/t.
\label{eq:lem:proof:1}
\end{equation}
Denote the event $A_i =\{|Q_{.i}^T\epsilon | \leq \lambda\}, i=1,2,\cdots,m$:
\begin{eqnarray*}
\{ \|Q^T\epsilon\|_\infty \leq \lambda\} &=& \{\max_{1\leq i \leq m} |Q_{.i}^T\epsilon | \leq \lambda \} \\
                                   &=& \cap_{i=1}^m A_i. 
\end{eqnarray*}
We derive the probability as follows:
\begin{eqnarray*}
\Pr(\cap_{i=1}^m A_i) &=& 1 - \Pr(\cup_{i=1}^m A_i^c) \\
                 &\geq& 1 - \sum_{i=1}^m \Pr(A_i^c) \\
                 &=& 1 - m\Pr(|\xi| \geq \lambda/\sigma)\\
                 &\geq& 1 - 2\frac{m\sigma}{\lambda}e^{-\lambda^2/2\sigma^2},
\end{eqnarray*}
where the second inequality holds by union bound and the last inequality follows \eqref{eq:lem:proof:1} since $\xi \sim \mathcal{N}(0,1)$.
Thus, we can take $\lambda = 2\sigma\sqrt{\log m}$ so that
\begin{eqnarray*}
\Pr(\|Q^T\epsilon)\|_\infty \leq \lambda) \geq 1 - 1/m\sqrt{\log m} =  1 - O(1/m).
\end{eqnarray*}
It completes the proof.
\end{proof}

\noindent{\bf RIP Condition Implies GR Condition}

Recall that to exactly recover $\beta^*$ the RIP condition requires that $\delta_{2s} > \sqrt{2}-1$ \citep{candes2008restricted}, where $s=\|\beta^*\|_0$ and $\delta_{2s}$ is defined as
for matrix $X^TX$ is defined as
\[
1-\delta_{2s} \leq {\|Xg\|^2 \over \|g\|^2} \leq 1+ \delta_{2s}\quad \forall~\|g\|_0 \leq 2s.
\]
Now we need to prove that $\delta_{2s} > \sqrt{2}-1$ leads to the GR condition $\rho(X,X, 2, \|\beta^*\|_0) >0$.

Consider an arbitrary vector $h$ and an arbitrary index set $T_0 \subset \{1,2,\cdots, m\}$ with cardinality $s$ satisfying $\|h_{T_0}\|_1 \geq \|h_{T_0^c}\|_1$. $T_1$ corresponds to the locations of the $s$ largest coefficients of $h_{T_0^c}$; $T_2$ to the locations of the next s largest coefficients of $h_{T_0^c}$, and so on. We use $T_{01}$ to denote $T_0\cup T_1$ for short.

Note the fact that if $Xh \neq 0$, then $X^TXh\neq 0$. To obtain $\rho(X,X, 2, \|\beta^*\|_0) >0$, it suffices to show that for any nonzero $h$ satisfying $\|h_{T_0^c}\|_1 \leq \|h_{T_0}\|_1$, the following ratio is positive
\begin{equation}
{\|Xh\| /\|h\|} > 0.
\label{eq:great0} 
\end{equation}
We have
\begin{equation*}
\begin{aligned}
 &|\langle Xh, Xh_{T_{01}} \rangle|
\\ \geq & \|Xh_{T_{01}}\|^2 - \sum_{j\geq 2} |\langle Xh_{T_{01}}, Xh_{T_j} \rangle|
\\ \geq & \|Xh_{T_{01}}\|^2 -\sum_{j\geq 2} |\langle Xh_{T_{0}}, Xh_{T_j} \rangle| -\sum_{j\geq 2} |\langle Xh_{T_{1}}, Xh_{T_j} \rangle|
\\ \geq & \|Xh_{T_{01}}\|^2 - \delta_{2s}\sum_{j\geq 2}(\|h_{T_0}\|+ \|h_{T_1}\|)\|h_{T_j}\|
\\ \geq & \|Xh_{T_{01}}\|^2 - \sqrt{2}\|h_{T_{01}}\|\delta_{2s}\sum_{j\geq 2}\|h_{T_j}\|
\\ \geq & \|Xh_{T_{01}}\|^2 - \sqrt{2}\|h_{T_{01}}\|\delta_{2s}\|h_{T_0^c}\|_1s^{-1/2}
\\ \geq & (1-\delta_{2s})\|h_{T_{01}}\|^2 - \sqrt{2}\delta_{2s}\|h_{T_{01}}\|^2
\\ \geq & (1-(\sqrt{2}+1)\delta_{2s})\|h_{T_{01}}\|^2,
\end{aligned}
\end{equation*}
where the third inequality uses the result $|\langle Xh_{T_i}, Xh_{T_j} \rangle| \leq \delta_{2s} \|h_{T_i}\|\|h_{T_j}\|$ if $i\neq j$, see Lemma 2.1 \citet{candes2008restricted}). It follows from the fact $|\langle Xh, Xh_{T_{01}} \rangle| \leq \|Xh\|\|Xh_{T_{01}}\|$ that 
\begin{equation}
 \|Xh\| \geq (1-(\sqrt{2}+1)\delta_{2s})\|h_{T_{01}}\|.
 \label{eq:Xh}
\end{equation}
From $\|h_{T_0^c}\|_1 \leq \|h_{T_0}\|_1$, we have $\|h\| \leq 2\|h_{T_{01}}\|$, see the last line of the proof for Theorem 1.2 in \citep{candes2008restricted}. It also can be found from \citep{candes07dantzig, bickel2009simultaneous}. Together with \eqref{eq:Xh} and the RIP condition $\delta_{2s}< \sqrt{2}-1$, we obtain
\[ 
\|Xh\| \geq (1-(\sqrt{2}+1)\delta_{2s})\|h_{T_{01}}\| \geq \frac{(1-(\sqrt{2}+1)\delta_{2s})}{2}\|h\|,
\]
which verifies \eqref{eq:great0}.

\noindent{\bf Comparison to Existing Error Bounds for DS}

This section aims to show that the error bound provided in \eqref{eq:bound} is a tighter bound than two existing results in \citep{candes07dantzig} and \citep{bickel2009simultaneous}. For simpler notations, we denote the difference between the estimate $\hat{\beta}_{\text{DS}}$ by DS and the true model $\beta^*$ as $h=\hat{\beta}_{\text{DS}} - \beta^*$, $T$ denotes the support set of $\beta^*$, and $s$ denotes the sparsity of $\beta^*$. The complete comparison requires extensive space to basically repeat the proofs in \citep{candes07dantzig} and \citep{bickel2009simultaneous}. Here, we just show their results, highlight the key point in their original proofs, and illustrate why our error bound does not loose theirs. 

Existing results conducted in \citep{candes07dantzig} and \citep{bickel2009simultaneous} are only based on the following facts
\begin{itemize}  [noitemsep,topsep=0pt,leftmargin=*]
\item (FACT 1) $\|h_{T^c}\|_1 \leq \|h_T\|_1$ (Please refer to Eq.~(3.2) in \citep{candes07dantzig} and Eq.~(B.12) in \citep{bickel2009simultaneous});
\item (FACT 2) $\|X^TXh \|_{\infty} \leq 2\|X^T\epsilon\|_{\infty} \leq_{(\text{P})} 2\sigma \sqrt{\log m}$ (Please refer to Eq.~(3.3) in \citep{candes07dantzig} and Eq.~(B.7) in \citep{bickel2009simultaneous}).
\end{itemize} 
which are also the foundations to provide our error bound in \eqref{eq:bound}. 

The error bound provided in Theorem 1.1 of \citep{candes07dantzig} is derived from (see the end of Proof of Theorem 1.1 in \citep{candes07dantzig})
\begin{equation}
\begin{aligned}
\|h\| \overset{(\text{FACT 1})}{\leq} & \frac{2\sqrt{s}}{1-\delta - \theta} \|X^TXh\|_\infty 
\\
\overset{(\text{FACT 2})}{\leq} & \frac{2\sqrt{s}}{1-\delta - \theta} 2\|X^T\epsilon\|_\infty.
\end{aligned}
\label{eq:candesbound}
\end{equation}
Please check the original paper for the definitions of $\delta$ and $\theta$. The error bound (with $p=2$, $m=s$) provided in \citep{bickel2009simultaneous} is derived from 
\begin{equation}
\begin{aligned}
\|h\|  \overset{(\text{FACT 1})}{\leq} & \frac{32 s}{\kappa(s, s, 1)}\|X^TX\epsilon\|_\infty 
\\
\overset{(\text{FACT 2})}{\leq} &
\frac{32 s}{\kappa(s, s, 1)} 2\|X^T\epsilon\|_\infty.
\end{aligned}
\label{eq:bickelbound}
\end{equation}
Please refer to the original paper for the definition of $\kappa(s, s, 1)$. 

From these two bounds in \eqref{eq:candesbound} and \eqref{eq:bickelbound}, they essentially uses FACT 1 to find an upper bound for $\|h\| / \|X^TXh\|_{\infty}$. This observation motivates us to directly define the upper bound through the definition of $\rho(\cdot, \cdot, \cdot)$ in \eqref{eq:kappa}
\[
\rho(X,X, s, 2) \leq \frac{\|X^TXh\|_\infty} {\|h\|} \quad \forall h~\text{under FACT 1}.  
\]
Then we simply apply FACT 2 as \eqref{eq:candesbound} and \eqref{eq:bickelbound} to obtain the error bound in \eqref{eq:bound}. Therefore, our analysis provides a tighter error bound than \citet{candes07dantzig} and \citet{bickel2009simultaneous}. 
}

{
\bibliographystyle{abbrvnat}
\bibliography{bo}

\begin{thebibliography}{31}
\providecommand{\natexlab}[1]{#1}
\providecommand{\url}[1]{\texttt{#1}}
\expandafter\ifx\csname urlstyle\endcsname\relax
  \providecommand{\doi}[1]{doi: #1}\else
  \providecommand{\doi}{doi: \begingroup \urlstyle{rm}\Url}\fi

\bibitem[Becker et~al.(2011)Becker, Cand{\`e}s, and Grant]{tfocs:becker2011}
S.~R. Becker, E.~J. Cand{\`e}s, and M.~C. Grant.
\newblock Templates for convex cone problems with applications to sparse signal
  recovery.
\newblock \emph{Mathematical Programming Computation}, 3\penalty0 (3):\penalty0
  165--218, 2011.

\bibitem[Bickel et~al.(2009)Bickel, Ritov, and
  Tsybakov]{bickel2009simultaneous}
P.~J. Bickel, Y.~Ritov, and A.~B. Tsybakov.
\newblock Simultaneous analysis of lasso and dantzig selector.
\newblock \emph{The Annals of Statistics}, pages 1705--1732, 2009.

\bibitem[Candes and Tao(2007)]{candes07dantzig}
E.~Candes and T.~Tao.
\newblock {The Dantzig selector: Statistical estimation when p is much larger
  than n}.
\newblock \emph{The Annals of Statistics}, pages 2313--2351, 2007.

\bibitem[Candes(2008)]{candes2008restricted}
E.~J. Candes.
\newblock The restricted isometry property and its implications for compressed
  sensing.
\newblock \emph{Comptes Rendus Mathematique}, 346\penalty0 (9):\penalty0
  589--592, 2008.

\bibitem[Candes and Tao(2005)]{rip:candes2005decoding}
E.~J. Candes and T.~Tao.
\newblock Decoding by linear programming.
\newblock \emph{IEEE Transactions on Information Theory}, 51\penalty0
  (12):\penalty0 4203--4215, 2005.

\bibitem[Candes et~al.(2008)Candes, Wakin, and Boyd]{WDS}
E.~J. Candes, M.~B. Wakin, and S.~P. Boyd.
\newblock Enhancing sparsity by reweighted l1 minimization.
\newblock \emph{Journal of Fourier analysis and applications}, 14\penalty0
  (5-6):\penalty0 877--905, 2008.

\bibitem[Chatterjee et~al.(2014)Chatterjee, Chen, and Banerjee]{ds:general}
S.~Chatterjee, S.~Chen, and A.~Banerjee.
\newblock Generalized dantzig selector: Application to the k-support norm.
\newblock In \emph{Advances in Neural Information Processing Systems}, pages
  1934--1942, 2014.

\bibitem[Chen and Donoho(1994)]{bp:chen1994basis}
S.~Chen and D.~Donoho.
\newblock Basis pursuit.
\newblock In \emph{IEEE the Twenty-Eighth Asilomar Conference onSignals,
  Systems and Computers}, volume~1, pages 41--44, 1994.

\bibitem[Chen et~al.(2001)Chen, Donoho, and Saunders]{bpdn:chen2001atomic}
S.~S. Chen, D.~L. Donoho, and M.~A. Saunders.
\newblock Atomic decomposition by basis pursuit.
\newblock \emph{SIAM review}, 43\penalty0 (1):\penalty0 129--159, 2001.

\bibitem[Cong et~al.(2011)Cong, Yuan, and Liu]{cong2011sparse}
Y.~Cong, J.~Yuan, and J.~Liu.
\newblock Sparse reconstruction cost for abnormal event detection.
\newblock In \emph{Computer Vision and Pattern Recognition (CVPR), 2011 IEEE
  Conference on}, pages 3449--3456. IEEE, 2011.

\bibitem[Donoho(2006)]{donoho2006compressed}
D.~L. Donoho.
\newblock Compressed sensing.
\newblock \emph{IEEE Transactions on Information Theory}, 52\penalty0
  (4):\penalty0 1289--1306, 2006.

\bibitem[Geist et~al.(2012)Geist, Scherrer, Lazaric, and
  Ghavamzadeh]{DantzigRL:2012}
M.~Geist, B.~Scherrer, A.~Lazaric, and M.~Ghavamzadeh.
\newblock {A Dantzig Selector Approach to Temporal Difference Learning}.
\newblock In \emph{International Conference on Machine Learning}, 2012.

\bibitem[Liu et~al.(2012)Liu, Mahadevan, and Liu]{ROTD:NIPS2012}
B.~Liu, S.~Mahadevan, and J.~Liu.
\newblock Regularized off-policy {TD}-learning.
\newblock In \emph{Advances in Neural Information Processing Systems}, pages
  845--853, 2012.

\bibitem[Liu et~al.(2010{\natexlab{a}})Liu, Zhang, Jiang, and
  Liu]{ds:group:liu2010group}
H.~Liu, J.~Zhang, X.~Jiang, and J.~Liu.
\newblock The group dantzig selector.
\newblock In \emph{International Conference on Artificial Intelligence and
  Statistics}, pages 461--468, 2010{\natexlab{a}}.

\bibitem[Liu(2014)]{liu1914statistical}
J.~Liu.
\newblock \emph{Statistical Methods for Genome-wide Association Studies and
  Personalized Medicine}.
\newblock PhD thesis, University of Wisconsin-Madison, 2014.

\bibitem[Liu et~al.(2010{\natexlab{b}})Liu, Wonka, and
  Ye]{ds:multi:liu2010multi}
J.~Liu, P.~Wonka, and J.~Ye.
\newblock Multi-stage dantzig selector.
\newblock In \emph{Advances in Neural Information Processing Systems}, pages
  1450--1458, 2010{\natexlab{b}}.

\bibitem[Liu et~al.(2013)Liu, Fujimaki, and Ye]{liu2013forward}
J.~Liu, R.~Fujimaki, and J.~Ye.
\newblock Forward-backward greedy algorithms for general convex smooth
  functions over a cardinality constraint.
\newblock \emph{ICML}, 2013.

\bibitem[Lu et~al.(2012)Lu, Pong, and Zhang]{ds:lu2012alternating}
Z.~Lu, T.~K. Pong, and Y.~Zhang.
\newblock An alternating direction method for finding dantzig selectors.
\newblock \emph{Computational Statistics \& Data Analysis}, 56\penalty0
  (12):\penalty0 4037--4046, 2012.

\bibitem[Mahadevan and Liu(2012)]{mahadevan2012sparse}
S.~Mahadevan and B.~Liu.
\newblock Sparse q-learning with mirror descent.
\newblock \emph{arXiv preprint arXiv:1210.4893}, 2012.

\bibitem[Nesterov(2004)]{nesterov2004introductory}
Y.~Nesterov.
\newblock \emph{Introductory lectures on convex optimization: A basic course},
  volume~87.
\newblock Springer, 2004.

\bibitem[Qin et~al.(2014)Qin, Li, and Janoos]{qin2014sparse}
Z.~Qin, W.~Li, and F.~Janoos.
\newblock Sparse reinforcement learning via convex optimization.
\newblock In \emph{Proceedings of the 31st International Conference on Machine
  Learning (ICML-14)}, pages 424--432, 2014.

\bibitem[Scherrer(2010)]{Scherrer:ObliqueProjection}
B.~Scherrer.
\newblock Should one compute the temporal difference fix point or minimize the
  bellman residual? the unified oblique projection view.
\newblock In \emph{ICML}, pages 52--68, 2010.

\bibitem[Tibshirani(1996)]{lasso:tibshirani1996regression}
R.~Tibshirani.
\newblock Regression shrinkage and selection via the lasso.
\newblock \emph{Journal of the Royal Statistical Society. Series B
  (Methodological)}, pages 267--288, 1996.

\bibitem[Tropp(2004)]{tropp2004greed}
J.~A. Tropp.
\newblock Greed is good: Algorithmic results for sparse approximation.
\newblock \emph{IEEE Transactions on Information Theory}, 50\penalty0
  (10):\penalty0 2231--2242, 2004.

\bibitem[Tsitsiklis and Van~Roy(1997)]{tsitsiklis-roy:tdfun}
J.~Tsitsiklis and B.~Van~Roy.
\newblock An analysis of temporal-difference learning with function
  approximation.
\newblock \emph{IEEE Transactions on Automatic Control}, 42:\penalty0 674--690,
  1997.

\bibitem[Van De~Geer et~al.(2009)Van De~Geer, B{\"u}hlmann,
  et~al.]{van2009conditions}
S.~A. Van De~Geer, P.~B{\"u}hlmann, et~al.
\newblock On the conditions used to prove oracle results for the lasso.
\newblock \emph{Electronic Journal of Statistics}, 3:\penalty0 1360--1392,
  2009.

\bibitem[Wang and Yuan(2012)]{ds:wang2012linearized}
X.~Wang and X.~Yuan.
\newblock The linearized alternating direction method of multipliers for
  dantzig selector.
\newblock \emph{SIAM Journal on Scientific Computing}, 34\penalty0
  (5):\penalty0 A2792--A2811, 2012.

\bibitem[Yu and Bertsekas(2010)]{yu2010error}
H.~Yu and D.~P. Bertsekas.
\newblock Error bounds for approximations from projected linear equations.
\newblock \emph{Mathematics of Operations Research}, 35\penalty0 (2):\penalty0
  306--329, 2010.

\bibitem[Zhang(2009)]{zhang2009consistency}
T.~Zhang.
\newblock On the consistency of feature selection using greedy least squares
  regression.
\newblock In \emph{Journal of Machine Learning Research}, pages 555--568, 2009.

\bibitem[Zhang(2011{\natexlab{a}})]{zhang2011adaptive}
T.~Zhang.
\newblock Adaptive forward-backward greedy algorithm for learning sparse
  representations.
\newblock \emph{IEEE Transactions on Information Theory}, 57\penalty0
  (7):\penalty0 4689--4708, 2011{\natexlab{a}}.

\bibitem[Zhang(2011{\natexlab{b}})]{zhang2011sparse}
T.~Zhang.
\newblock Sparse recovery with orthogonal matching pursuit under rip.
\newblock \emph{IEEE Transactions on Information Theory}, 57\penalty0
  (9):\penalty0 6215--6221, 2011{\natexlab{b}}.

\end{thebibliography}
}

\end{document}